\pdfoutput=1

\documentclass[11pt]{article}

\usepackage{EMNLP2023}

\usepackage{times}
\usepackage{latexsym}
\usepackage[T1]{fontenc}
\usepackage[utf8]{inputenc}
\usepackage{microtype}
\usepackage{inconsolata}
\usepackage{graphicx}
\usepackage{tabularx}
\usepackage{booktabs, multirow} 
\usepackage{makecell}

\usepackage[justification=centering]{caption}
\usepackage{comment}
\usepackage[shortlabels]{enumitem}
\usepackage{caption}
\renewcommand{\footnotesize}{\fontsize{7pt}{11pt}\selectfont}

\newcommand{\BUSTER}{\textit{BUSTER}}
\newcommand{\BUSTERSP}{\textit{BUSTER} }
\newcommand{\TITLE}{\BUSTER: a ``BUSiness Transaction Entity Recognition'' dataset }
\newcommand{\FORMa}{\textit{Form 8K}}
\newcommand{\FORMb}{\textit{8-K}}
\newcommand{\EXBa}[0]{\textit{Exhibit 99.1}}
\newcommand{\EXBb}[0]{\textit{EX-99.1}}
\newcommand{\NSPRINTS}{$8$ }
\newcommand{\SPRINTSIZE}{$500$ }
\newcommand{\SPRINTSHARED}{$100$ }
\newcommand{\SPRINTSHAREDPERC}{$20\%$}
\newcommand{\TOTALSIZE}{$4000$ }
\newcommand{\TOTALSHARED}{$800$ }

\newcommand{\togroup}[1]{\textit{#1}}
\newcommand{\Parties}{\togroup{Parties }}
\newcommand{\Advisors}{\togroup{Advisors }}
\newcommand{\Info}{\togroup{Generic\_Info }}

\newcommand{\tolabel}[1]{\textit{#1}}
\newcommand{\ACQUIREDCOMPANY}{\tolabel{ACQUIRED\_COMPANY }}
\newcommand{\BUYINGCOMPANY}{\tolabel{BUYING\_COMPANY }}
\newcommand{\SELLINGCOMPANY}{\tolabel{SELLING\_COMPANY }}

\newcommand{\LEGALCONSULTINGCOMPANY}{\tolabel{LEGAL\_CONSULTING\_COMPANY }}
\newcommand{\GENERICCONSULTINGCOMPANY}{\tolabel{GENERIC\_CONSULTING\_COMPANY }}

\newcommand{\ANNUALREVENUES}{\tolabel{ANNUAL\_REVENUES }}

\newcommand{\nump}[1]{\#(#1)}
\newcommand{\nums}[1]{\#(#1)}
\newcommand{\ann}[1]{\mbox{\textit{ann}$_{#1}$}}
\newcommand{\kin}[0]{k \in \{1,2\}}
\newcommand{\N}[0]{N=\nump{L_1 \cup L_2}}
\newcommand{\tagset}{tag-set}
\newcommand{\Tagset}{Tag-set}
\newcommand{\fold}[1]{\mbox{\textit{fold}$_{#1}$}}

\newcommand{\DATAPOINTS}[0]{tags}
\newcommand{\OverallAgreementDegree}[0]{{Joint Probability of Agreement}}
\newcommand{\AgrSym}[2]{\mbox{\textit{JPA}}}
\newcommand{\AgrForm}[2]{\frac{\nump{L_{#1} \cap L_{#2}}}{\nump{L_{#1} \cup L_{#2}}}}
\newcommand{\AgrEq}[2]{\AgrSym{#1}{#2} = \AgrForm{#1}{#2}}
\newcommand{\MutualAgreementDegree}[0]{{Conditional Probability of Agreement}}
\newcommand{\MutAgrSymT}[3]{\mbox{\textit{CPA}}_{#1}}
\newcommand{\MutAgrFormT}[3]{\frac{\nump{L_{#2} \cap L_{#3}}}{\nums{L_{#1}}}}
\newcommand{\MutAgrEqT}[3]{\MutAgrSymT{#1}{#2}{#3} = \MutAgrFormT{#1}{#2}{#3}}
\newcommand{\Coverage}[0]{{Coverage}}
\newcommand{\CovSymT}[3]{\mbox{\textit{Cov}}_{#1}}
\newcommand{\CovFormT}[3]{\frac{\nums{L_{#1}}}{\nump{L_{#2} \cup L_{#3}}}}
\newcommand{\CovEqT}[3]{\CovSymT{#1}{#2}{#3} = \CovFormT{#1}{#2}{#3}}
\newcommand{\CohenK}[0]{{Cohen's kappa ($\kappa$)}}
\newcommand{\CohenKSym}[0]{\kappa}

\newcommand{\CohenKFormT}[2]{\frac{p_o - p_e}{1 - p_e}}
\newcommand{\CohenKEqT}[2]{\CohenKSym = \CohenKFormT{#1}{#2}}
\newcommand{\po}[0]{p_o = \AgrSym {1}{2}}

\newcommand{\peT}{p_e = \frac{\nums{L_1} \times \nums{L_2}}{N^2}}

\newcommand{\az}[1]{\color{black} #1}
\newcommand{\lr}[1]{\color{black} #1}
\title{\TITLE}
\author{
  Andrea Zugarini\\
  expert.ai, Siena, Italy\\
  \texttt{azugarini@expert.ai}\\
  \And
  Andrew Zamai\\
  expert.ai, Siena, Italy\\
  \texttt{azamai@expert.ai}\\
  \AND
  Marco Ernandes\\
  expert.ai, Siena, Italy\\
  \texttt{mernandes@expert.ai}
  \And
  Leonardo Rigutini \\
  expert.ai, Siena, Italy\\
  \texttt{lrigutini@expert.ai} \\
}
\begin{document}
\maketitle
\begin{abstract}
Albeit Natural Language Processing has seen major breakthroughs in the last few years, transferring such advances into real-world business cases can be challenging. 
One of the reasons resides in the displacement between popular benchmarks and actual data.
Lack of supervision, unbalanced classes, noisy data and long documents often affect real problems in vertical domains such as finance, law and health.
To support industry-oriented research, we present \BUSTER, a BUSiness Transaction Entity Recognition dataset. 
The dataset consists of 3779 manually annotated documents on financial transactions. 
We establish several baselines exploiting both general-purpose and domain-specific language models. The best performing model is also used to automatically annotate 6196 documents, which we release as an additional silver corpus to \BUSTER.
\end{abstract}
%
%
%
\begin{table*}[ht!]
    \centering
    \small
    \begin{tabularx}{\textwidth}{crll}
        \toprule
        \textbf{Tag Family} &\textbf{Tag Name} &\textbf{Description} \\
        \midrule
        \multirow{5}{*}{\Parties}
                    & \BUYINGCOMPANY     & The company which is acquiring the target.\\ \\
                    & \SELLINGCOMPANY    & The company which is selling the target.\\ \\
                    & \ACQUIREDCOMPANY   & The company target of the transaction.\\
        \midrule
        \multirow{5}{*}{\Advisors}
                    & \LEGALCONSULTINGCOMPANY & \makecell{A law firm providing advice on the transaction, such as:\\ government regulation, litigation, anti-trust,\\ structured finance, tax etc.}\\ \\
                    & \GENERICCONSULTINGCOMPANY & \makecell{A general firm providing any other type of advice,\\ such as: financial, accountability, due diligence, etc.}\\
        \midrule
        \multirow{1}{*}{\Info}
                    & \ANNUALREVENUES    & \makecell{The past or present annual revenues of\\ any company or asset involved in the transaction.}\\
        \bottomrule
    \end{tabularx}
    \caption{Description of the \tagset \ defined in \BUSTER.}
    \label{tab:tagset}
\end{table*}

\section{Introduction}
\label{sec:introduction}
Natural Language Processing (NLP) is a field potentially beneficial to a broad span of language-intensive domains, such as law and health. Whilst lots of Financial data are tabular, there is also crucial information stored in reports, news, transaction agreements, etc. 

The abrupt developments in NLP~\citep{vaswani2017attention} are favouring its adoption as assistance tools for human experts in many tasks, ranging from Document Classification~\citep{chalkidis-etal-2019-large} to Information Extraction~\citep{alvarado2015domain,loukas2022finer} and even Text Summarization~\citep{bhattacharya2019comparative}. However, transferring the emerging technologies into industry applications can be non-trivial. Adapting Large Language Models (LLMs) to vertical domains usually requires fine-tuning on domain-specific annotated data. Labeling is often a time-consuming, expensive process, especially when experts in the field are involved. 
Recently, several benchmarks and datasets have been
constructed for law~\citep{chalkidis-etal-2022-lexglue}, health~\citep{li2016biocreative} and finance~\citep{loukas2022finer}.

\begin{figure}[ht]
    \centering
    \includegraphics[width=1.0\columnwidth]{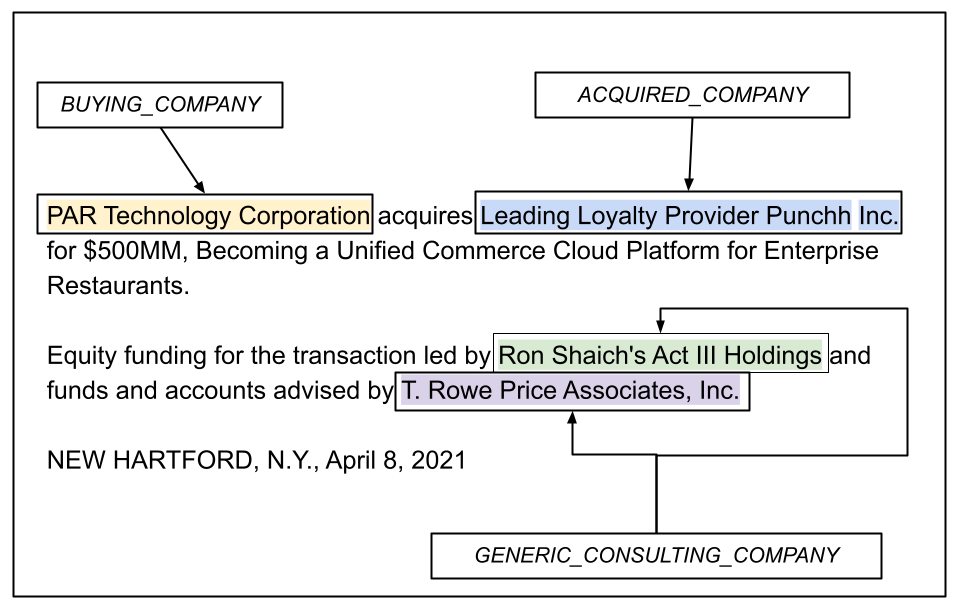}
    \caption{An annotated example extracted from \BUSTER.}
    \label{fig:buster_doc_example}
\end{figure}

In this work, we support industry-oriented research community by presenting \BUSTER: a BUSiness Transaction Entity Recognition dataset. 
{\lr As the title suggests, \BUSTER \ is an Entity Recognition (ER) benchmark that focus on the main actors involved in a business transaction. After collecting about ten thousands business transaction documents from EDGAR company acquisition reports, we constructed a dataset with 3779 manually annotated documents (the Gold corpus), from which we trained an LLM to automatically annotate the remaining 6196 documents (the Silver corpus). We analyze the properties of the proposed dataset and also evaluate the performance of some baselines.}
The dataset will be public and free to download as a benchmark for the NLP community. 

The paper is organized as follows. First we review in Section~\ref{sec:related_works} previous related works on Financial NER and document-level datasets. Then, we describe the data collection process and annotation methodologies in sections \ref{sec:data_collection} and \ref{subsec:annotation}, respectively.
A detailed description of \BUSTER \ and its statistics follows in Section~\ref{sec:buster}. In Section~\ref{sec:experiments} we establish baselines with different LLMs. Finally, in Section~\ref{sec:conclusions} we draw our conclusions and outline possible future research directions. 
%
\section{Related works}
\label{sec:related_works}
%
%
{\lr Several document datasets in the financial domain have been proposed in the literature, but few of them are dedicated to the Entity Recognition (ER) task. Furthermore, these few are mainly intended for the standard Named Entity Recognition (NER) task, such as~\citep{alvarado2015domain,francis2019transfer,hampton2016automated,kumar-etal-2016-experiments}. }
{\lr In \citet{alvarado2015domain} is presented a corpus (FIN) of eight documents from SEC which were manually annotated using the standard four NER data type: person, organization, location and miscellaneous.
Unlike that dataset, in \BUSTERSP we decided to focus on Entities that are involved in a financial transaction.}
{\lr FiNER-139~\cite{loukas2022finer} instead consists in a large corpus of SEC documents annotated via gold XBLR tags, that includes a label set of 139 numerical entities on about 1.1M sentences. The tag attribution mostly depends on context rather than the token itself, as it is in \BUSTER. Beside the completely different tag set, the main difference between \BUSTERSP and Finer-139 is the fact that we release a document-level benchmark. Indeed, the detection of roles like the buyer company can require scopes wider than a single sentence. Moreover, documents come from files with heterogeneous layouts, extensions and structure, which can sometimes hinder the segmentation of the document into single sentences.}

{\lr Outside the financial domain, a variety of document-level datasets for NER have been proposed.}
{\lr DocRED~\citep{yao2019docred} is a NER and Relation Extraction (RE) corpus built from Wikidata and Wikipedia short text passages, while BioCreative~\cite{li2016biocreative} is  a dataset for NER/RE on health domain.}
{\lr In ~\citep{quirk2016distant}, the authors propose a dataset for NER in medical area.}
%
%
\begin{table*}[ht!]
\centering
\small
\begin{tabularx}{\textwidth}{cr|c|cc|cc|c}
    \toprule
    \textbf{} & \textbf{} & $\AgrSym{1}{2}$ & $\MutAgrSymT{1}{2}{1}$ & $\MutAgrSymT{2}{1}{2}$ & $\CovSymT{1}{2}{1}$ & $\CovSymT{2}{1}{2}$ & $\CohenKSym$ \\
    \midrule
    \multirow{3}{*}{\Parties}
                        & \BUYINGCOMPANY                & 0.6514 & 0.7445 & 0.8389 & 0.8749 & 0.7764 & 0.6810 \\
                        & \SELLINGCOMPANY               & 0.5026 & 0.6362 & 0.7053 & 0.7900 & 0.7126 & 0.6383 \\
                        & \ACQUIREDCOMPANY              & 0.5611 & 0.6658 & 0.7811 & 0.8427 & 0.7184 & 0.6119 \\
    \midrule
    \multirow{2}{*}{\Advisors}
                        & \LEGALCONSULTINGCOMPANY       & 0.8913 & 0.9011 & 0.9880 & 0.9891 & 0.9022 & 0.9405 \\
                        & \GENERICCONSULTINGCOMPANY     & 0.6624 & 0.7273 & 0.8814 & 0.9108 & 0.7516 & 0.7862 \\
    \midrule
    \multirow{1}{*}{\Info}
                        & \ANNUALREVENUES               & 0.5781 & 0.6894 & 0.7817 & 0.7590 & 0.7000 & 0.7246 \\
    \toprule
                        & \textbf{MICRO OVERALL}        & 0.6100 & 0.7107 & 0.8115 & 0.8583 & 0.7517 & 0.7257 \\
                        & \textbf{MACRO OVERALL}        & 0.6448 & 0.7504 & 0.8148 & 0.8566 & 0.7882 & 0.7402 \\
    \bottomrule
\end{tabularx}
\caption{The quality assessment results of the output of the annotation process.}
\label{tab:buster_internal_assessment}
\end{table*}
\section{Data Collection}
\label{sec:data_collection}
    Our goal was to create a highly business-oriented dataset to recognize relevant entities involved in financial transactions. Unlike standard NER tasks, we focused on the problem of entity-role recognition, where the goal is to identify a set of entities but only where they appear with specific roles in a context, such as companies involved in an acquisition or consultants assisting in an operation.
    %
\subsection*{Target documents}
\label{sec:target_documents}
    To collect such documents, we exploited the EDGAR (Electronic  Data  Gathering,  Analysis,  and  Retrieval system) service of the U.S. Securities and Exchange Commission (SEC)~\footnote{\url{https://www.sec.gov/edgar/}}. 
    The SEC’s mission is to maintain fair, orderly, and efficient markets. In particular, the organization aims to give transparency to business activities and provide investors with more security on the companies in which they invest, facilitating capital formation. For this purpose, domestic and foreign companies conducting business in the US are required to provide regular reports to the SEC through EDGAR. Reports are filed based on a list  of  forms  that  correspond  to  certain  filing  types. 
    %
    The EDGAR service provides more than 150 different form types (\textit{filing type})~\footnote{\url{https://en.wikipedia.org/wiki/SEC_filing}} and of these, the \FORMa\ type deserves particular attention. 
    
    An \FORMb\ provides investors with timely notification of significant changes at listed companies such as acquisitions, bankruptcy, the resignation of directors, or changes in the fiscal year~\footnote{\url{https://www.sec.gov/investor/pubs/readan8k.pdf}}. 
    Optionally, but very frequently, the \FORMa\ includes a document called \EXBa\ (often abbreviated on \EXBb). It consists of a disclosure document which summarizes all the details of the operation announced in the form and it is designed to provide investors with a complete and detailed view on the operation. 
    %
\subsection*{Crawling, filtering and processing}
\label{sec:crawling_etc}
    To collect the \EXBb\ disclosure documents from EDGAR reporting company acquisitions, ownership changes and share purchase, we make use of the full index tool of the EDGAR site. Limiting to 2021, we downloaded about $120,000$ \EXBb\ disclosure documents in HTML format.
    After parsing, cleaning and removing any empty or too short documents, we selected the relevant documents using transaction-related keywords (acquisition, acquire, ownership, etc.) obtaining a final raw dataset of about $10,000$ text files.
%
\section{Annotation}
\label{subsec:annotation}
    For data labeling, we used a double-blind manual procedure. 
    Specifically, we utilized two annotators (\ann{1} and \ann{2}), who were trained on the financial transactions topic and who were provided with a \tagset \ and specific guidelines to follow in the entity tagging procedure.
    The annotation procedure has been performed using the expert.ai natural language platform.
    It consists in an integrated environment for deep language understanding and provides a complete natural language workflow with end-to-end support for annotation, labeling, model training, testing and workflow orchestration~\footnote{\url{https://www.expert.ai/products/expert-ai-platform/}}.
\subsection*{\Tagset}
\label{sec:tagset}
    In designing the \tagset, we identified three families of \DATAPOINTS: (a) \Parties which groups \DATAPOINTS \ used to identify the entities directly involved in the transaction; (b) \Advisors which groups \DATAPOINTS \ identifying any external facilitator and advisor of the transaction and (c) \Info which identifies \DATAPOINTS \ reporting any information about the transaction.
    For each family, we defined a set of related \DATAPOINTS.
    The \tagset \ is reported in Table \ref{tab:tagset}.
\subsection*{Guidelines and General instructions}
\label{sec:guidelines}
    In order to improve annotation coherency, the schema definitions outlined in Table \ref{tab:tagset} were prepared as guidelines to the annotators. 
    Moreover, the following general instructions were provided:
    \begin{itemize}
        \item \textbf{Annotate linguistically apparent instances only} -- Tag only instances of entities where the class is linguistically evident. Do not tag a string just because you know that it is an instance of an entity: the context must make it obvious that it is an instance of such class.
        \item \textbf{Evaluate sentence context only} -- Tag only instances of entities in which there is evidence within a sentence that the instance is of that entity. Each sentence should be evaluated for entities in isolation from the rest of the document context. 
    \end{itemize}
%
%
%
\begin{table*}[ht!]
    \centering
    \small
    \begin{tabular}{cr||ccccc|c||c}
        \toprule
        &                             & \multicolumn{6}{c||}{\textbf{Gold}}          & \textbf{Silver}  \\
        &                             & \fold{1}  & \fold{2}  & \fold{3}  & \fold{4} & \fold{5} & \textit{Total} & \textit{Total}\\
        \midrule
        & \textbf{N. Docs}            & 753       & 759       & 758       &  755     & 754  & 3779  & 6196   \\
        & \textbf{N. Tokens}          & 685K     &  680K      & 687K      &  697K    & 688K & 3437K & 5647K  \\
        & \textbf{N. Annotations}     & 4119      & 4267      & 4100      &  4103    & 4163 & 20752 & 33272  \\
        \midrule
        \multirow{4}{*}{\Parties}
        & \BUYINGCOMPANY              & 1734      & 1800      & 1721      &  1707    & 1717 & 8679  & 14558 \\
        & \SELLINGCOMPANY             &  460      &  447      &  456      &  426     & 439  & 2228  & 4016  \\
        & \ACQUIREDCOMPANY            & 1399      & 1473      & 1362      &  1430    & 1447 & 7111  & 9879  \\
        & \textbf{Total}              & 3593      & 3720      & 3539      &  3563    & 3603 & 18018 & 28453 \\
        \midrule
        \multirow{3}{*}{\Advisors}
        & \LEGALCONSULTINGCOMPANY     & 142       & 132       & 152       &   146    & 153  & 721   & 1176  \\
        & \GENERICCONSULTINGCOMPANY   & 256       & 267       & 261       &   248    & 256  & 1279  & 2210  \\
        & \textbf{Total}              & 398       & 399       & 413       &   394    & 409  & 2013  & 3545  \\
        \midrule
        \multirow{2}{*}{\Info} 
        & \ANNUALREVENUES             & 128       & 148       & 148       &   146    & 151  & 721   & 1274  \\
        & \textbf{Total}              & 128       & 148       & 148       &   146    & 151  & 696   & 1274  \\
        \bottomrule
    \end{tabular}
    \caption{The statistics of the 5 folds Gold and Silver data.}
    \label{tab:fold_stats}
\end{table*}
\subsection*{Annotation Procedure}
\label{sec:procedure}
    To monitor the annotation procedure, the data set was divided into ``sprints'' which have been provided sequentially to the annotators.
    Each sprint consists of a pair of document batches that have been submitted independently to the two annotators.
    Additionally, we designed each sprint so that its two batches shared a certain percentage of documents.
    In this way, in each sprint, a portion of documents will be tagged by both annotators. 
    Although this choice reduces the number of documents processed over time, it allows subsequent estimation of the annotation quality in each sprint.
    
    We set the size of each sprint to \SPRINTSIZE documents, \SPRINTSHARED of which were shared between the two annotators (\SPRINTSHAREDPERC). The two annotators processed \NSPRINTS sprints, thus obtaining \TOTALSIZE annotated documents, \TOTALSHARED of which were labeled by both annotators. 
    Finally, after removing documents without any labels, the resulting dataset was composed of $3779$ labeled documents.
%
\subsection*{Validation}
\label{validation}
    To evaluate the quality output of the annotation process, we exploited the shared set of documents that had been tagged by both annotators.
    In particular, indicating with $L_1$ and $L_2$ the two sets of annotations~\footnote{Each `annotation' refers to an entire annotated phrase.} inserted respectively by the two annotators \ann{1} and \ann{2} in the shared documents, we calculated several standard indexes~\footnote{\url{https://en.wikipedia.org/wiki/Inter-rater_reliability}}:
    \begin{enumerate}[label=(\alph*)]
        %
        \item \OverallAgreementDegree, which measures the chance of having a match between the two annotators: \mbox{$\AgrEq{1}{2}$}.
        \item \MutualAgreementDegree \ of \ann{k}, which measures the naive probability that annotations inserted by an annotator $k$ have a match with annotations entered by the other: \mbox{$\MutAgrEqT{k}{1}{2}$}, \mbox{$\kin$}.
        \item \Coverage \ of \ann{k}, which measures the probability that a randomly selected annotation was entered by the annotator $k$: \mbox{$\CovEqT{k}{1}{2}$}, \mbox{$\kin$}.
        \item \CohenK, which extends the \OverallAgreementDegree \ taking into account that agreement may occur by chance \cite{Cohen1960ACO}: \mbox{$\CohenKEqT{1}{2}$} where \mbox{$\po$} is the observed agreement, \mbox{$\peT$} \ estimates the probability of a random agreement and $\N$ is the total number of inserted annotations.
    \end{enumerate}
    The results are reported in the Table \ref{tab:buster_internal_assessment} and the values of \CohenK \ show a substantial agreement between the two evaluators~\cite{landis1977measurement}.
\subsection*{Managing annotations in shared documents}
\label{sec:managing_annotations}
    In creating the final dataset, it was required to manage shared sets annotated by both annotators. 
    Firstly, we accepted all non-overlapping annotations from both annotators. 
    Secondly, we fixed overlapping, incoherent, annotations by involving a third annotator who manually assigned the correct label.
    Moreover, for pairs of overlapping annotations with boundaries $l_1=[s_1,e_1]$ and $l_2=[s_2,e_2]$, we merged them into a new annotation such that \mbox{$l = [s,e]=[\min(s_1,s_2),\max(e_1,e_2)]$}.
%
\begin{table*}[ht!]
\centering
\small
\begin{tabular}{r|ccc|ccc}
    \toprule
    \textbf{Model}  &  $\mu$-\textbf{Precision} &  $\mu$-\textbf{Recall} & $\mu$-\textbf{F1} & M-\textbf{Precision} & M-\textbf{Recall} & M-\textbf{F1}\\ 
    \midrule
        BERT        & 61.16 $\pm$ 1.65 & 67.42 $\pm$ 2.72 & 64.06 $\pm$ 0.90 & 55.12 $\pm$ 1.75 & 66.60 $\pm$ 2.79 & 59.80 $\pm$ 1.23\\
        
        SEC-BERT    & 66.76 $\pm$ 0.74 & 74.18 $\pm$ 1.99 & 70.28 $\pm$ 0.90 & 70.30 $\pm$ 0.96 & 78.10 $\pm$ 1.82 & 73.98 $\pm$ 1.14\\
        
        \textbf{RoBERTa}     & \textbf{69.84 $\pm$ 1.41} & \textbf{75.08 $\pm$ 1.42} & \textbf{72.34 $\pm$ 0.39} & \textbf{72.38 $\pm$ 0.64} & \textbf{79.34 $\pm$ 1.17} & \textbf{75.58 $\pm$ 0.66}\\
        
        Longformer  & 69.28 $\pm$ 2.71 & 73.40 $\pm$ 1.31 & 71.24 $\pm$ 1.34 & 70.02 $\pm$ 3.27 & 77.34 $\pm$ 1.49 & 73.30 $\pm$ 2.25\\
    \bottomrule
    \end{tabular}\caption{Micro ($\mu$-) and macro (M-) scores of the four baseline models evaluated using 5-Fold Cross Validation.}\label{tab:baselines_micro-f1}
\end{table*}
\section{The \BUSTER \ dataset}
\label{sec:buster}
    The final \BUSTER \ dataset is composed of $3779$ labeled documents.    
    In Figure~\ref{fig:buster_doc_example}, we show an example of an annotated text passage inside a document.
    As explained, those documents were manually annotated and represent the ``gold'' \BUSTER \ corpus.
    We randomly split the data into 5 folds to yield a  statistically robust benchmark.
    Indeed, such division allows the use of a standard \mbox{k-fold} \mbox{cross-validation} approach.
    
    The data set has been used as benchmark for $4$ state-of-the art ER models (as described in Section~\ref{sec:experiments}) and the best performing model has been used to automatically annotate the remaining $6196$ documents.
    The resulting annotated data was released as a ``silver'' extra corpus in \BUSTER \ benchmark.
    The details of the 5 folds and of the silver extra corpus are reported in Table~\ref{tab:fold_stats}.\\\\
    The full \BUSTER \ benchmark is publicly available and free-to-download from the expert.ai website\footnote{\url{https://www.expert.ai/buster}} and on HuggingFace\footnote{\url{https://huggingface.co/datasets/expertai/BUSTER}} and we are confident that it can become a point of reference in the field of Entity Recognition, in particular for the financial sector.
\subsection*{Statistics}
\begin{figure}[!h]
        \centering
        \includegraphics[width=0.9\columnwidth,trim=0 0 0 40,clip]{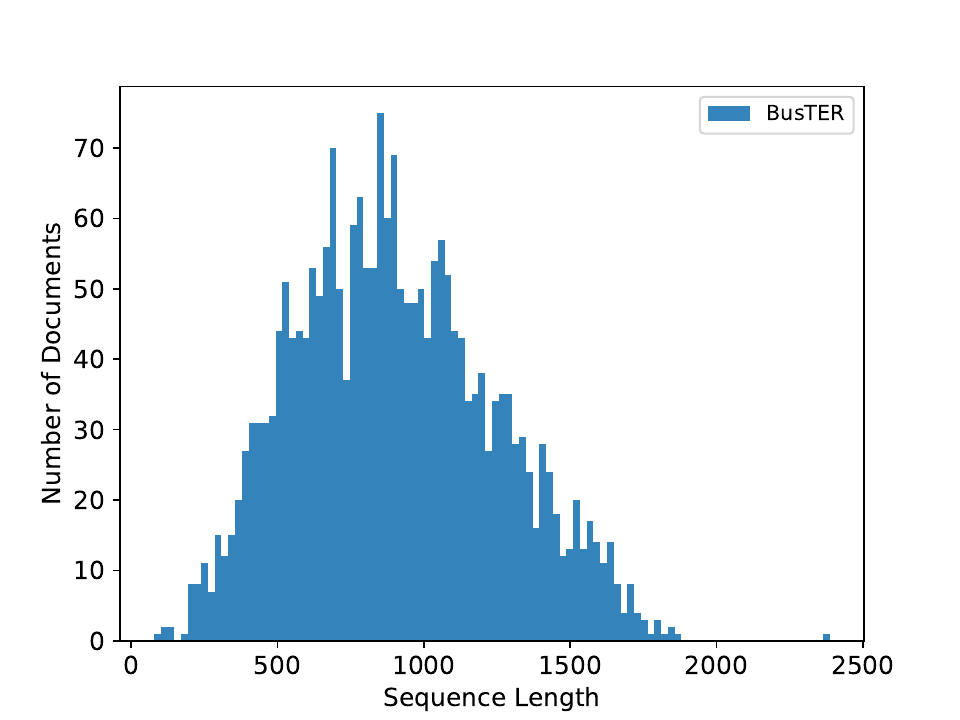}
        \caption{Sequence length distribution of \BUSTER \ documents in terms of words.}
        \label{fig:seq_len_distr}
    \end{figure}

    Figure \ref{fig:seq_len_distr} shows the distribution of document lengths. 
    The documents appear to have an average length of around 700 words and most of them fall into the 500-1000 range.
    Also, documents with more than 2000 words are extremely rare.

\begin{figure}[!h]
        \centering
        \includegraphics[width=0.9\columnwidth,trim=0 0 0 40,clip]{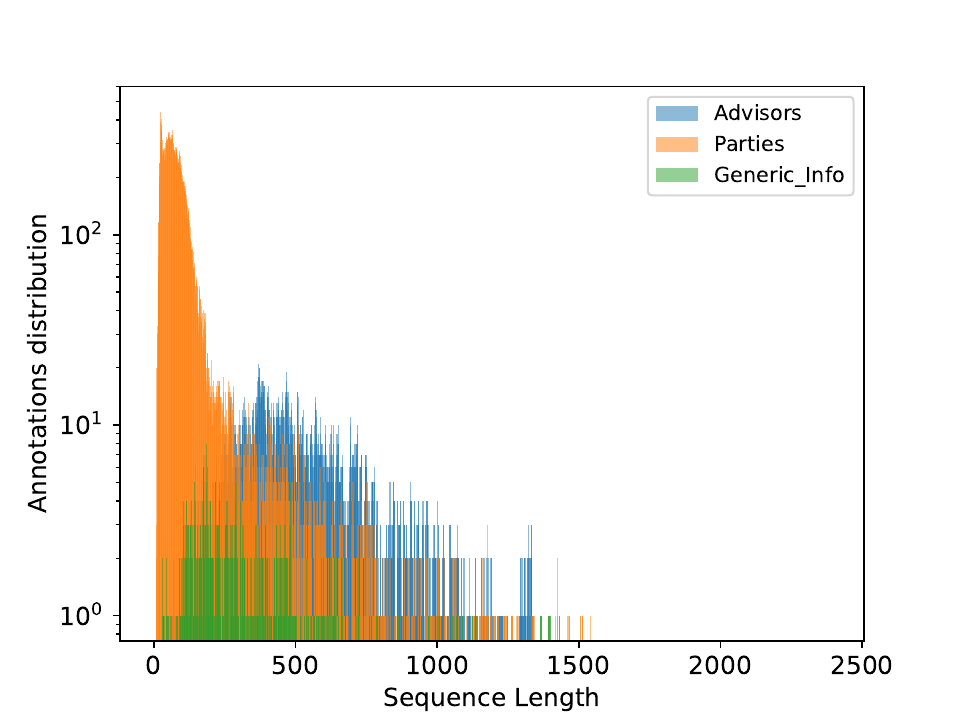}
        \caption{Distribution of \DATAPOINTS \ families inside the documents.}
        \label{fig:perclass_distribution}
    \end{figure}

%
%
%
    In figure \ref{fig:perclass_distribution}, we report the distribution of the three \DATAPOINTS \ families based on their position within the documents. 
    We can observe how the \DATAPOINTS \ belonging to the \Parties family (in orange) are centered in the initial parts of the documents, while the remaining are distributed more uniformly and, in any case, located towards the second part of documents. 
    However, no \DATAPOINTS \ occurs beyond the 1500th word.
%
%
\begin{table*}[ht!]
    \centering
    \small
    \begin{tabular}{cr|ccc}
     \toprule
        & \textbf{} &\textbf{Precision} & \textbf{Recall} & \textbf{F1}\\ 
     \midrule
     \multirow{3}{*}{\Parties}
        & \BUYINGCOMPANY   & 74.06 $\pm$ 2.06 & 78.38 $\pm$ 1.47 & 76.12 $\pm$ 0.85\\
        & \SELLINGCOMPANY  & 65.34 $\pm$ 2.35 & 75.04 $\pm$ 3.15 & 69.82 $\pm$ 0.77\\
        & \ACQUIREDCOMPANY & 64.42 $\pm$ 1.11 & 70.38 $\pm$ 0.63 & 67.26 $\pm$ 0.38\\
    \midrule
    \multirow{2}{*}{\Advisors}
        & \LEGALCONSULTINGCOMPANY   & 84.86 $\pm$  3.33 & 90.90 $\pm$  2.33 & 87.72 $\pm$ 1.46\\
        & \GENERICCONSULTINGCOMPANY & 73.98 $\pm$  1.97 & 77.98 $\pm$  3.27 & 75.90 $\pm$ 2.04\\
    \midrule
    \multirow{1}{*}{\Info}
        & \ANNUALREVENUES & 61.88 $\pm$ 5.95 & 79.36 $\pm$ 4.66 & 69.30 $\pm$ 4.24\\
    \bottomrule
    \end{tabular}
    \caption{Tag-wise precision, recall and F1-score values obtained with\\ the RoBERTa baseline using 5-Fold Cross Validation.}
    \label{tab:roberta_cls_report}
\end{table*}
\section{Experiments}
\label{sec:experiments}
To establish baselines, we performed several experiments using both generic and domain-specific language models.
%
%
\subsubsection*{Experimental Setup}
\label{sec:expertimental_setup}
{\az In the experiments, we followed a 5-folds cross validation approach using the folds in Table~\ref{tab:fold_stats} .}
%
\paragraph{Metrics.} We adopt traditional NER metrics for evaluation, i.e. micro and macro F1 scores, referred as $\mu$-F1 and M-F1, respectively. True positives are counted in a strict sense, i.e. an entity is considered correctly predicted if and only if all of its constituent tokens are well identified, and no additional tokens belong to the entity.

\paragraph{Dealing with long documents.} 
As shown in Figure~\ref{fig:seq_len_distr}, the vast majority of documents in \BUSTER \ has more than 500 words, which typically exceeds the maximum sequence length that LLMs (e.g. BERT \cite{devlin2018bert}) can take in input. Truncation would cause a major drop of most of the document and a significant loss of information. Therefore, we split documents into contiguous chunks of text. Chunking is done such that no token is truncated at all and we fill each chunk sequence as much as possible. All the baselines are trained and tested on chunks with the exception of Longformers, since they are capable of processing longer sequences up to 4096 tokens.
\subsection*{Baseline Models}
\label{sec:baseline_models}
We considered several transformer-based models that report state-of-the-art performance in NLP. In particular, we have selected the following 4 models. 
\paragraph{BERT.} BERT~\cite{devlin2018bert} constitutes a standard baseline since it is one of the most popular LLMs nowadays.
\paragraph{RoBERTa.} Similarly to BERT, RoBERTa~\cite{liu2019roberta} is a widely-used Language Model in the NLP community. The model is an optimized version of BERT and generally outperforms it.
\paragraph{SEC-BERT.} We also consider a domain-specific LLM. We consider SEC-BERT~\cite{loukas2022finer}, a model pre-trained from scratch on EDGAR-CORPUS, a large collection of financial documents~\cite{loukas2021edgar}.
\paragraph{Longformer.} Longformer~\cite{beltagy2020longformer} is a transformer architecture equipped with self-attention mechanisms that scales linearly with the sequence length. Longformers were specifically designed to deal with long documents, hence they are a natural good candidate for processing \BUSTER.
%
\subsection*{Results}
\label{sec:results}
The baselines' performance are presented in Table~\ref{tab:baselines_micro-f1}.
{\lr RoBERTa turned out to be the best performing model, with Longformer achieving similar levels of accuracy. BERT base, instead, under-performed with respect to the other baselines. However, when fine-tuning BERT on the financial domain (SEC-BERT) there is a clear F1 improvement.}

{\lr Inspecting the scores of single entity tags obtained by the best model, i.e. RoBERTa (Table~\ref{tab:roberta_cls_report}), we can observe that the \Advisors family is generally well captured by the model.}
{\lr For \Parties and \Info families instead, the results are different.}
{\lr The model performs very well on \BUYINGCOMPANY, while \ACQUIREDCOMPANY, \SELLINGCOMPANY and \ANNUALREVENUES appear more complex to discriminate, especially in terms of precision.}
{\lr In our analysis, this depends on some structural characteristics of these entities. The first two tags (\ACQUIREDCOMPANY and \SELLINGCOMPANY) are strongly related to each other and often they are not easy to disambiguate even for human annotators, as confirmed by the quality assessment outlined in Table~\ref{tab:buster_internal_assessment}.}
{\lr The definition of \ANNUALREVENUES instead, is very specific and detailed (Section~\ref{sec:guidelines}) and this makes it hard to distinguish it from occurrences of other economic data present in the text, e.g. EBITDA.}
{\lr Finally, the inherent complexity inevitably increases the noise in the gold annotations, thus affecting the training of the model itself.}
\section{Conclusions and future works}\label{sec:conclusions}
{\lr In this work, we presented \BUSTER, an Entity Recognition (ER) benchmark for business transaction-related entities. It consists of a corpus of 3779 manually annotated documents on financial transaction (the Gold data) which has been randomly divided into 5 folds, plus an additional set of 6196 automatically annotated documents (the Silver data) that were created from the fine-tuned RoBERTa model.}

The full \BUSTER \ benchmark is publicly available and free-to-download from the expert.ai website\footnote{\url{https://www.expert.ai/buster}} and on HuggingFace\footnote{\url{https://huggingface.co/datasets/expertai/BUSTER}} and we are confident that it can become a point of reference in the field of Entity Recognition, in particular for the financial sector.

{\lr In the future, we intend to work in two directions. On one side, we plan to increase the amount of manually labeled data and to extend the labels set with more transaction-related tags. On the other hand, we aim to introduce some specific types of relations between entities in order to extend the dataset to Relational Extraction.}
\subsection*{Acknowledgements}
A huge thank you to Bianca Vallarano and Stefano Genua who participated as annotators.
Thanks to Daniela Baiamonte who supported us in the production of the guidelines and in the validation of the annotation process.
Thanks to Paolo Lombardi who prepared the scripts to download and process the documents from EDGAR.

This work was supported by the IBRIDAI project, a project financed by the Regional Operational Program “FESR 2014-2020” of Emilia Romagna (Italy), resolution of the Regional Council n. 863/2021.
\bibliography{custom}
\bibliographystyle{acl_natbib}
\onecolumn
\appendix


\end{document}